\documentclass[conference]{IEEEtran}
\IEEEoverridecommandlockouts
\usepackage{cite}
\usepackage{amsmath,amssymb,amsfonts}
\usepackage{algorithmic}
\usepackage{graphicx}
\usepackage{textcomp}
\usepackage{xcolor}
\usepackage{hyperref}
\def\BibTeX{{\rm B\kern-.05em{\sc i\kern-.025em b}\kern-.08em
    T\kern-.1667em\lower.7ex\hbox{E}\kern-.125emX}}
\begin{document}

	\begin{titlepage}
	\begin{center}
		
		\Huge
		\textbf{Adaptive Anchor Pairs Selection \\ in a TDOA-based System Through\\ Robot Localization Error Minimization			
		}
		
		\vspace{0.5cm}
		\LARGE
		Accepted version
		
		\vspace{1.5cm}
		
		\text{Marcin Kolakowski}
		
		\vspace{.5cm}
		\Large
		Institute of Radioelectronics and Multimedia Technology
		
		Warsaw University of Technology
		
		Warsaw, Poland,
		
		contact: marcin.kolakowski@pw.edu.pl

		\vspace{2cm}

	\end{center}
	
	\Large
	\noindent
	\textbf{Originally presented at:}
	
	\noindent
2021 Signal Processing Symposium (SPSympo), Lodz, Poland, 2021
	
	\vspace{.5cm}
	\noindent
	\textbf{Please cite this manuscript as:}
	
	\noindent
M. Kolakowski, "Adaptive Anchor Pairs Selection in a TDOA-based System Through Robot Localization Error Minimization," 2021 Signal Processing Symposium (SPSympo), LODZ, Poland, 2021, pp. 128-132, doi: 10.1109/SPSympo51155.2020.9593477
	
	\vspace{.5cm}
	\noindent
	\textbf{Full version available at:}
	
	\noindent
	\url{https://doi.org/10.1109/SPSympo51155.2020.9593477}

	%		\vspace{.5cm}
	%		\noindent
	%		\textbf{Additional information:}
	%		
	%		\noindent
	%		The dataset used in the study is available at Zenodo:
	%		
	%		\noindent
	%		Marcin Kolakowski. (2021). UWB Channel Impulse Responses Registered in a Furnished Apartment (Version 1.0) [Data set]. Zenodo. \url{http://doi.org/10.5281/zenodo.4742391}
	
	\vfill
	
	\large
	\noindent
	© 2021 IEEE. Personal use of this material is permitted. Permission from IEEE must be obtained for all other uses, in any current or future media, including reprinting/republishing this material for advertising or promotional purposes, creating new collective works, for resale or redistribution to servers or lists, or reuse of any copyrighted component of this work in other works.
\end{titlepage}

\title{Adaptive Anchor Pairs Selection \\ in a TDOA-based System Through\\ Robot Localization Error Minimization\\
\thanks{The research has been partially funded by the National Centre for Research
and Development, Poland under Grant AAL2/2/INCARE/2018.}}

%\title{Adaptive Anchor Pairs Selection \\ in a TDOA-based System Through\\ Robot Localization Error Minimization\\
%
%\thanks{
%}

\author{\IEEEauthorblockN{Marcin Kolakowski}
\IEEEauthorblockA{\textit{Warsaw University of Technology} \\
\textit{Institute of Radioelectronics and Multimedia Technology}\\
Warsaw, Poland\\
m.kolakowski@ire.pw.edu.pl}
}

\maketitle

\begin{abstract}
The following paper presents an adaptive anchor pairs selection method for ultra-wideband (UWB) Time Difference of Arrival (TDOA) based positioning systems. The method divides the area covered by the system into several zones and assigns them anchor pair sets. The pair sets are determined during calibration based on localization root mean square error (RMSE). The calibration assumes driving a mobile platform equipped with a LiDAR sensor and a UWB tag through the specified zones. The robot is localized separately based on a large set of different TDOA pairs and using a LiDAR, which acts as the reference. For each zone, the TDOA pairs set for which the registered RMSE is lowest is selected and used for localization in the routine system work. 
The proposed method has been tested with simulations and experiments. The results for both simulated static and experimental dynamic scenarios have proven that the adaptive selection of the anchor nodes leads to an increase in localization accuracy. In the experiment, the median trajectory error for a moving person localization was at a level of 25 cm.
\end{abstract}


\section{Introduction}
The ultra-wideband-based localization systems are among the most accurate solutions for indoor positioning. Under the right conditions they allow to localize objects with errors below 50 centimeters \cite{mendoza-silvaMetaReviewIndoorPositioning2019}. However, such positioning quality is rarely attainable as the accuracy depends on several issues, many of which are hard to control.

The main factor deteriorating the localization accuracy is Non-Line of Sight (NLOS) propagation, in which obstacles obscure the direct path between the localized device and the system infrastructure. Such working conditions are common in cramped interiors, e.g., office spaces, where the signals are delayed and attenuated by walls and pieces of furniture. The errors resulting from working under NLOS conditions vary but are usually above 1 m, especially when times of arrival are measured based on reflected signal components \cite{djaja-joskoNewMapBased2017}.

The problems resulting from NLOS propagation are especially visible in the case of person localization \cite{otimFDTDEmpiricalExploration2019}, where the user's body is an additional obstacle obscuring the visibility of system infrastructure for a large portion of time.

Additionally, the accuracy of the system depends on the geometrical configuration of the system infrastructure \cite{fanDistributedAnchorNode2018}. The analysis performed in \cite{kauneAccuracyAnalysisTDOA} has shown that for TDOA-based localization, the Cramer Rao Lower Bound (CRLB) varies for different spatial configurations of the localized tag and the anchors.

A careful selection of anchors (or anchor pairs in differential methods such as TDOA) used in the localization process may partially reduce the negative effects of the above factors.
One way to improve localization accuracy is to detect the anchors working in NLOS conditions and remove them from location computation. An example of such a method is presented in \cite{jiang-ningNonlineofsightMitigationTechnique2008}, where the results obtained under NLOS conditions are identified and removed from position calculation. Unfortunately, this approach is implementable only when the number of installed anchors is high, and at least three Line of Sight (LOS) operating anchors are available most of the time.

The second group of methods for anchor selection is based on CRLB analysis \cite{daiNearlyOptimalSensor2020}. The performed studies have shown that the CRLB varies with locations within the system operation area \cite{kauneAccuracyAnalysisTDOA} and differs for particular anchor sets. Unfortunately, the results of the CRLB analysis might be hard to implement as the CRLB is fully applicable only for  Line of Sight (LOS)  operation scenarios, which are rarely the case in real indoor environments.

Another approach to anchor selection is to use the anchors, which are closest to the tag. The selection might be performed based on propagation time measurements, or signal strength \cite{cantonpaternaBluetoothLowEnergy2017a}. Such methods assume that the chances of the direct path being obscured are low in the case of closely spaced devices. Additionally, high signal levels allow performing more accurate measurements.

The last notable anchor selection algorithms group is based on localization error minimization \cite{feilongOptimumReferenceNode2015,monicaUWBbasedLocalizationLarge2015}. In these methods, the tag is localized based on different sets of anchors (or anchor pairs in TDOA scenarios), and then the most accurate results are chosen.

In the paper, an anchor pairs selection method adopting a similar approach is presented. It is intended for indoor TDOA-based localization, but it might be used in other systems with slight modifications. The method divides the system operation area into zones and chooses the most favorable TDOA pairs for each of them based on the calibration performed using a mobile robot. One of the algorithms' advantage is that it does not require ensuring LOS working conditions or deploying a system with extensive infrastructure.

The paper is structured as follows. Section \ref{sec:concept} contains the description of the method. The results of the simulations and experiments are presented in Sections \ref{sec:simulations} and \ref{sec:experiments}, respectively. Section \ref{sec:conclusions} concludes the paper.

\section{Localization with anchor pairs selection}
\label{sec:concept}

\subsection{System calibration}
\begin{figure}[bp]
\centerline{\includegraphics[width=.9\linewidth]{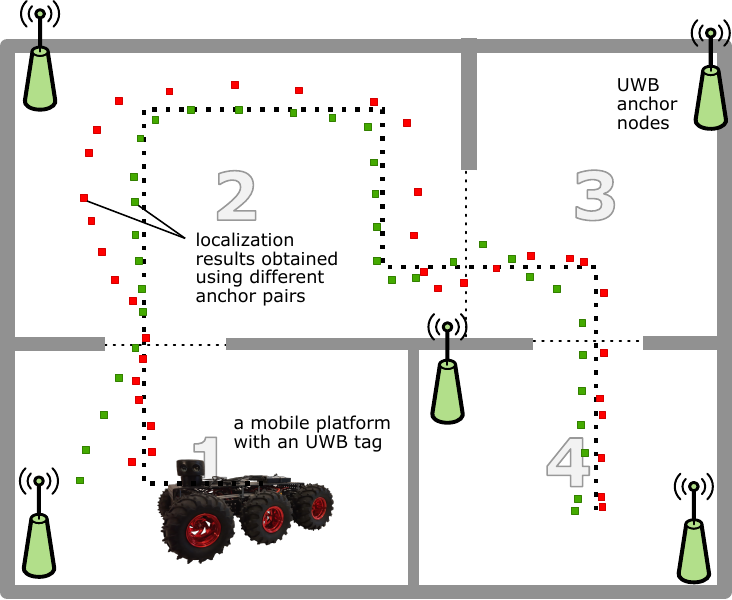}}
\caption{The proposed system calibration method concept.}
\label{fig:concept}
\end{figure}

The concept of the proposed anchor pairs selection method is illustrated with Fig. \ref{fig:concept}.

The proposed method assumes dividing the area covered by the system into separate zones and choosing the most favorable anchor pairs sets for each of them based on a calibration routine. The calibration requires using a robotic platform equipped with a LiDAR (Light Detection and Ranging) sensor and a UWB system tag. It consists of the following steps:
\begin{enumerate}
\item driving the platform through all zones while registering times of arrival of packets sent by the attached tags and taking LiDAR scans,
\item computing reference platform locations based on the LiDAR data,
\item estimating robot locations based on different sets of TDOA pairs,
\item choosing the best TDOA pairs set for each zone based on positioning RMSE values.
\end{enumerate}

The first step of the calibration procedure is driving the platform throughout the area where the system is installed. During the ride, the robot takes the LiDAR scans of the surroundings and the infrastructure measures time of arrival (TOA) of packets transmitted by the tag. The robot path should cover all of the zones - the higher the extent to which the areas are covered, the better the calibration result.

After the test drive, the reference robot locations are calculated based on the collected LiDAR scans. In the proposed algorithm, the robot is localized using a method based on occupancy grid maps. If the system deployment area map is unavailable, the environment should be mapped prior to the calibration, or a simultaneous localization and mapping (SLAM) algorithm should be used.

The robot location is then estimated based on different combinations of anchor pairs. In order to be effective, the algorithm should analyze as many TDOA pairs sets as possible. In the presented implementation, the list of analyzed sets is created as follows. First, a list of all possible, unique anchor pairs is created. Then the algorithm makes all possible combinations of these pairs (all unique combinations of three pairs, four pairs...). The obtained list of combinations might be filtered and differentiated among the zones to reduce the computational time of the algorithm. For example, combinations including far away anchors, which are too far to receive signals from the whole zone, might be safely removed as they would not be useful during the system's operation.

The robot location is computed using a Least Squares based estimator implemented using the Levenberg Maqrquadt method and minimizing the following expression: 
\begin{equation}
\min_{x} \sum_{i=0}^{n-1}(\rm{TDOA}_i^m -  \rm{TDOA}_i(x))^2
\label{eq:il_min}
\end{equation}
where $x$ is the sought robot location,  $n$ is the number of the TDOA pairs in the analyzed set and  $\rm{TDOA}_i^m$ and $\rm{TDOA}_i(x)$ are measured and estimated TDOA values.

The results obtained using different anchor pairs are assessed in each of the zones separately by calculating the root mean square error (RMSE). The combination of pairs, for which the RMSE is the lowest in the given zone, is selected and used later on by the localization algorithm in the system routine operation.

\subsection{Localization algorithm}
\label{sec:algorithm}
\begin{figure}[b]
\centerline{\includegraphics[width=0.5\linewidth]{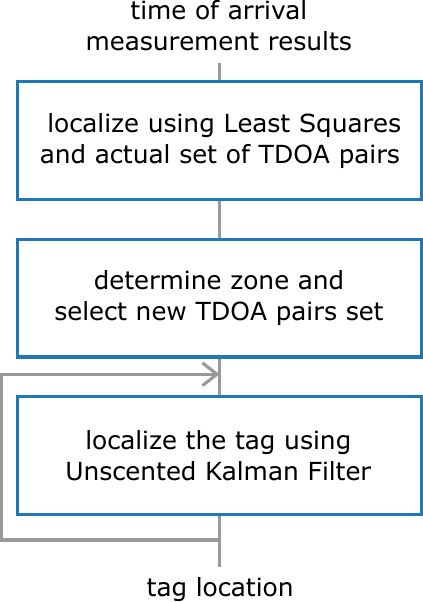}}
\caption{The workflow of the proposed localization algorithm.}
\label{fig:algorithm}
\end{figure}

The localization algorithm workflow consists of three steps, which are presented in Fig. \ref{fig:algorithm}.

The tag is first roughly localized using a Least Squares based estimator using the anchor pairs set from the previous algorithm iteration. Then the anchor pairs set used for localization is switched to the one corresponding to the detected zone. The final positioning is performed using an Unscented Kalman Filter based algorithm \cite{kolakowskiUWBBLETracking2020}. The result is used as the input in the next iteration.

\section{Simulations}
\label{sec:simulations}
\begin{figure}[htbp]
\centerline{\includegraphics[width=\linewidth]{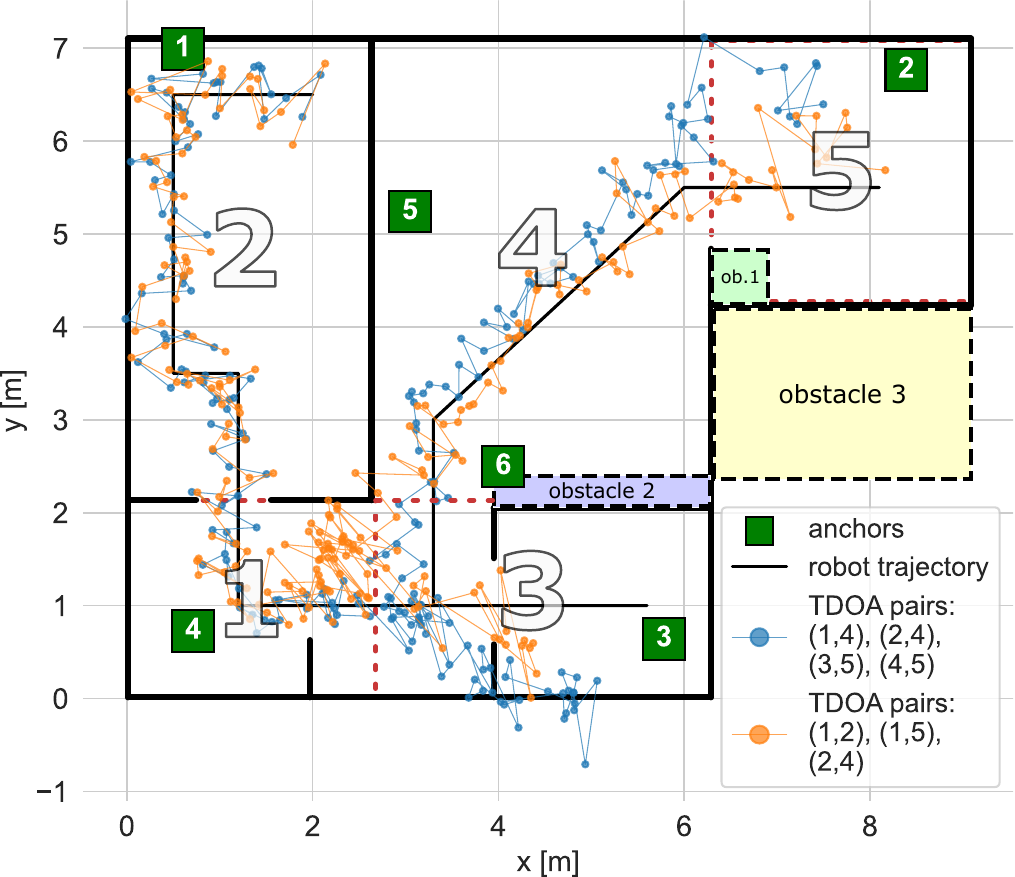}}
\caption{The layout of the simulated environment. The numbers specify the zones into which the apartment was divided. The presented localization results were obtained based on different anchor pairs sets.}
\label{fig:sim_env}
\end{figure}

The proposed method has been tested with simulations, which consisted of two steps: simulated system calibration and localization of static tags placed among the apartment. The simulations were performed using Python-based scripts for an environment presented in Fig. \ref{fig:sim_env}.

The simulated apartment consisted of two rooms, a corridor and a bathroom, which were divided into five zones marked on a plan. The deployed system comprised six anchors. The test path of the robot was simulated as several straight segments covering all of the rooms.

The simulator took into account the delays introduced by the walls (0.2 ns per traversed wall) and major obstacles (2, 2, 4 ns for obstacles 1, 2, 3, respectively). It was assumed that the measured time values were disturbed with a gaussian noise with a standard deviation of 0.6 ns. The LiDAR computed locations were estimated with a Gaussian error of 3 cm standard deviation.

The TDOA based locations were calculated for all possible sets of a minimum of three TDOA pairs, in which there were no more than three TDOA pairs including the same anchor. In total, 547 sets of anchor pairs were analyzed. The best anchor pairs for each zone are presented in Table \ref{tab:best_pairs}.

\begin{table}[htbp]
\caption{Most favorable sets of TDOA pairs (simulation)}
\begin{center}
\def\arraystretch{1.5}
\begin{tabular}{c|c}

\textbf{zone}&\textbf{anchor pairs}\\
\hline
1&(1, 2), (1, 4),  (2, 3),  (3, 4),  (4, 5)\\
2&(1, 4), (2, 4), (3, 5), (4, 5)\\
3&(1, 3), (1, 5), (2, 5), (3, 4), (4, 5)\\
4&(1, 2), (1, 4), (2, 4), (2, 5)\\
5&(1, 2), (1, 5), (2, 4)\\
\end{tabular}
\label{tab:best_pairs}
\end{center}
\end{table}

\begin{figure}[htbp]
\centerline{\includegraphics[width=\linewidth]{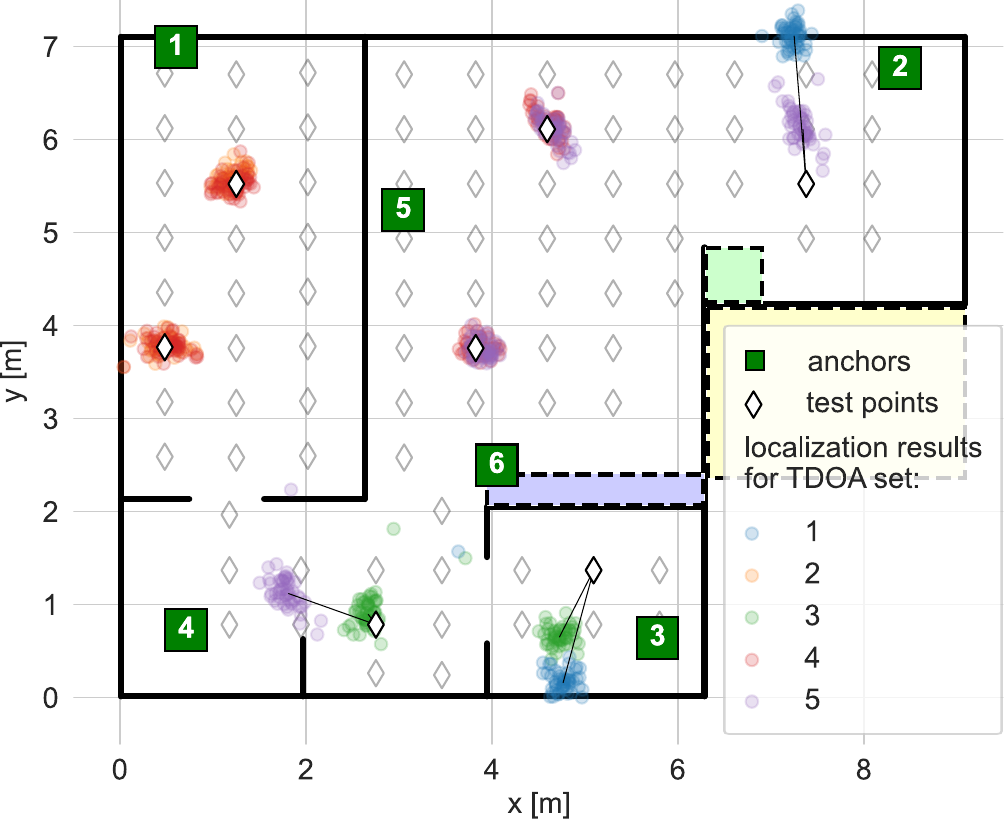}}
\caption{Exemplary simulation results.  Anchor pairs sets correspond to the sets for zones listed in Table \ref{tab:best_pairs}.}
\label{fig:sim_res}
\end{figure}

\begin{figure}[htbp]
\centerline{\includegraphics[width=\linewidth]{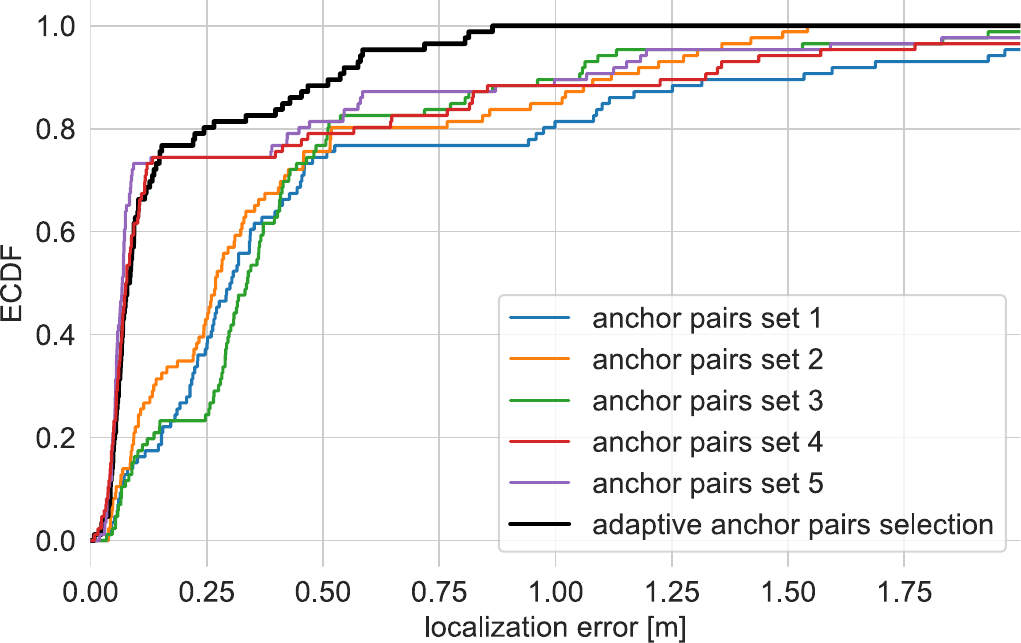}}
\caption{Estimated cumulative distribution function of mean localization errors. Anchor pairs sets correspond to the sets for zones listed in Table \ref{tab:best_pairs}.}
\label{fig:sim_cdfs}
\end{figure}

The effectiveness of the performed calibration was assessed based on the localization of static tags placed in 87 points distributed among the apartment. The simulation parameters were the same as in the case of the calibration phase. The tags were localized based on five anchor pairs sets listed in Table \ref{tab:best_pairs}. The layout of the test points and exemplary results are presented in Fig. \ref{fig:sim_res}. The Estimated Cumulative Distribution Function (ECDF) of mean localization error is presented in Fig. \ref{fig:sim_cdfs}.

The localization accuracy is different for the tested anchor pairs sets. When localizing the tags using the same sets of TDOAs in the whole apartment, the maximum mean localization error and its 80th percentile are always higher than 1.5 m and 45 cm, respectively. Using the proposed adaptive anchor pairs selection method, those values are much lower and are about 80 cm and 30 cm, respectively.

\section{Experiments}
\label{sec:experiments}

\begin{figure}[bp]
\centerline{\includegraphics[width=\linewidth]{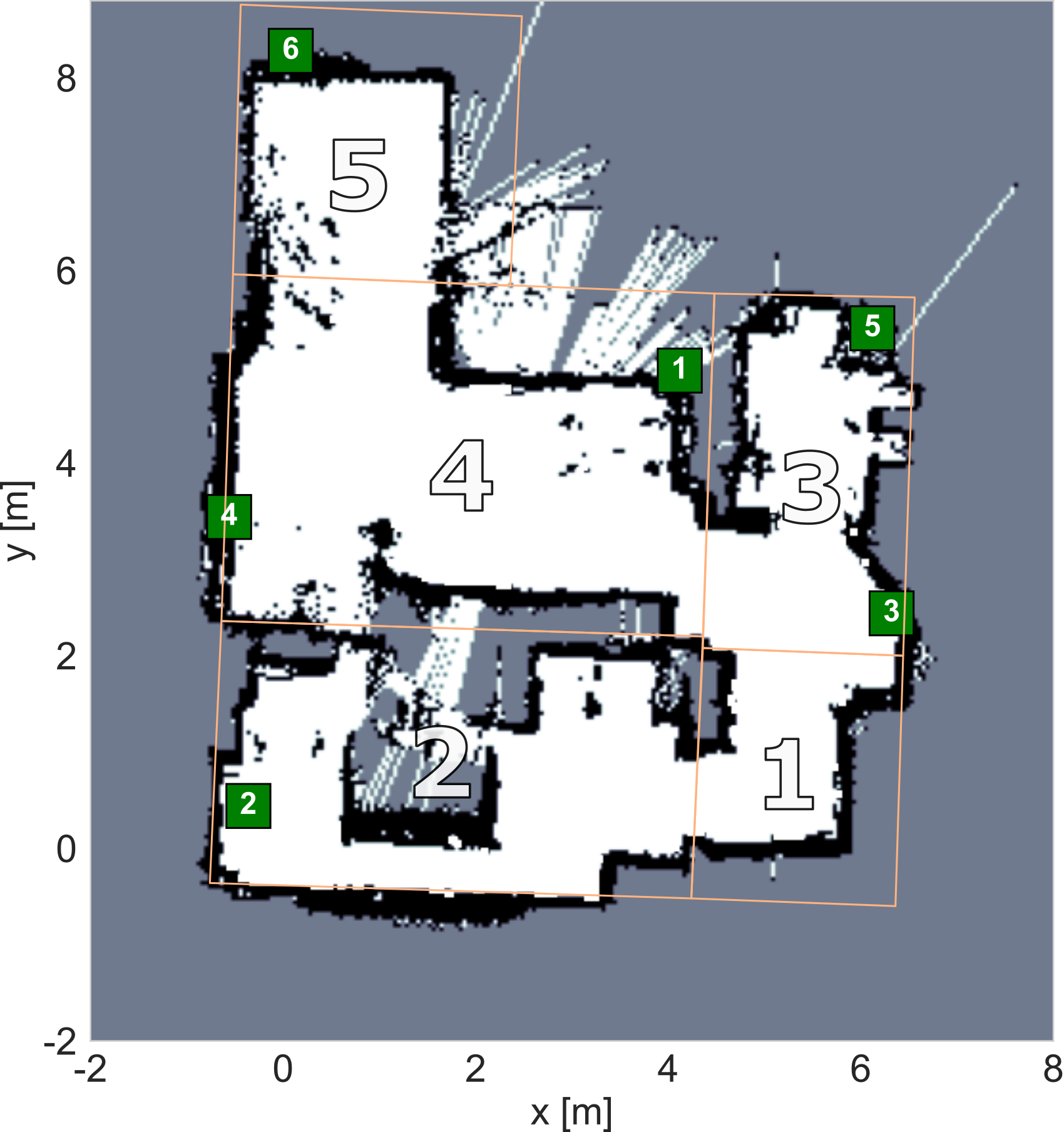}}
\caption{The map of the apartment created using the mobile platform. Anchor locations are marked with green squares}
\label{fig:exp_layout}
\end{figure}

The proposed method was tested with experiments in an apartment being a real-life version of the simulated one. The localization system used in the experiment was a hybrid UWB-BLE (Bluetooth Low Energy) positioning system \cite{kolakowskiUWBBLETracking2020} consisting of 6 anchors and one tag. During the study, only the UWB-part was used. The layout of the apartment and the placement of the system anchors is presented in Fig. \ref{fig:exp_layout}.

The mobile platform used for the system calibration was Dagu Wild Thumper 6WD All-Terrain Chassis with a Scanse Sweep LiDAR (Light Detection and Ranging) sensor and a Raspberry Pi-based controller. The tag was attached to the platform on a vertical wooden pole at 140 cm height. The LiDAR's field of view was at about 30 cm.

The experiment consisted of three phases:
\begin{enumerate}
\item mapping the environment and dividing it into zones,
\item determining the most favorable TDOA pairs for each zone,
\item localization of a moving person.
\end{enumerate}

In the first phase, the environment was mapped using the mobile platform. The mapping consisted in driving the platform through all of the apartment rooms registering scans using the attached LiDAR. The scans were then matched and the complete map of the apartment was created using the Iterative Closest Point and GraphSLAM algorithms. The resulting occupancy grid map of the apartment is presented in Fig.\ref{fig:exp_layout}. The apartment has been divided into zones similar to those assumed during the simulations.

\begin{figure}[htbp]
\centerline{\includegraphics[width=\linewidth]{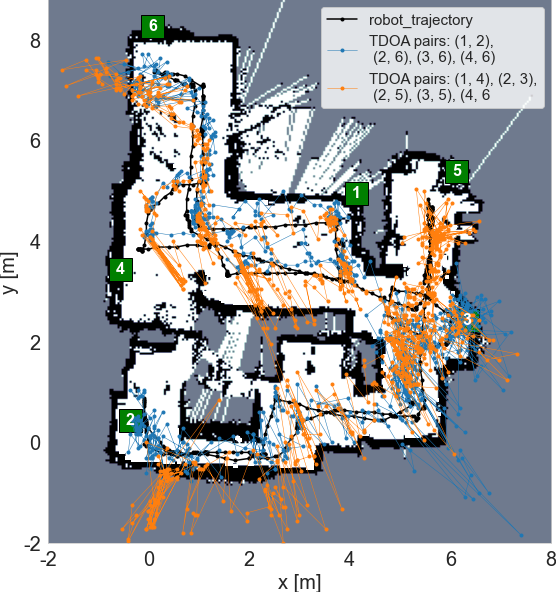}}
\caption{Movement trajectory of the mobile platform and its locations derived using the UWB system during the system calibration phase}
\label{fig:calib}
\end{figure}

The most favorable TDOA pairs sets for each of the zones were determined by driving the platform, once again, through the apartment and localizing it using the TDOA values calculated using different anchor pairs. In the study a total of 4487 TDOA pairs were analyzed (combinations of 3,4,5 pairs assuming that each anchor can be only used thrice in the given pairs set). The most favorable TDOA pairs for each of the zones are listed in Table \ref{tab:best_pairs_exp}. The calibration path of the robot and the exemplary localization results are presented in Fig.\ref{fig:calib}.

\begin{table}[htbp]
\caption{Most favorable sets of TDOA pairs (experiments)}
\begin{center}
\def\arraystretch{1.5}
\begin{tabular}{c|c}

\textbf{zone}&\textbf{anchor pairs}\\
\hline
1&(2,5), (3, 5),  (4,6)\\
2&(1, 2), (2, 6), (3, 6), (4, 6)\\
3&(1, 4), (2, 3), (2, 5), (3, 5), (4, 6)\\
4&(1, 2), (1, 3), (1, 5), (3, 5), (4, 5)\\
5&(1, 4), (2, 4), (2, 6), (3, 4)\\
\end{tabular}
\label{tab:best_pairs_exp}
\end{center}
\end{table}

\begin{figure}[t]
\centerline{\includegraphics[width=\linewidth]{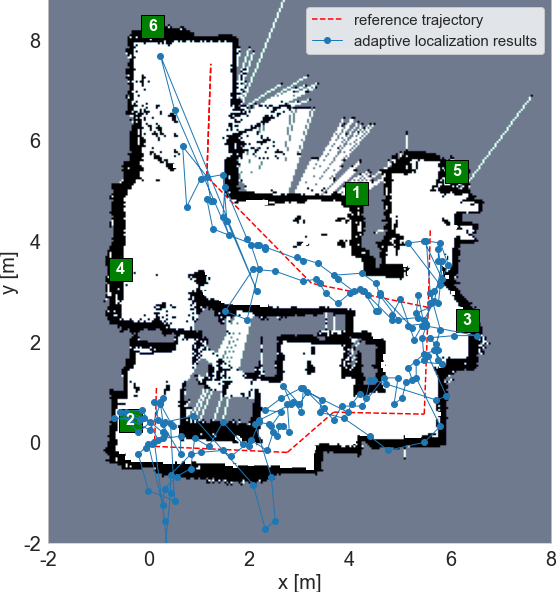}}
\caption{Moving person localization results}
\label{fig:results}
\end{figure}

\begin{figure}[htbp]
\centerline{\includegraphics[width=\linewidth]{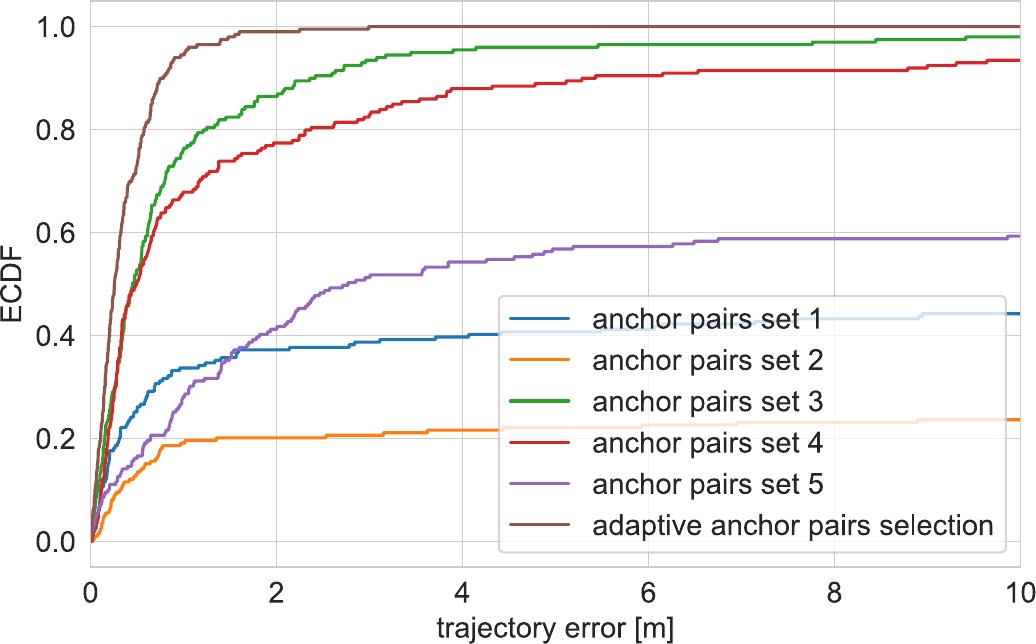}}
\caption{Empirical Cumulative Distribution Functions of trajectory error for different TDOA pairs sets}
\label{fig:cdfs}
\end{figure}

The last part of the experiment consisted in verifying the method's effectiveness by localizing a moving person using the algorithm proposed in section \ref{sec:algorithm}. The test trajectory and the localization results are presented in Fig.\ref{fig:results}. The Empirical Cumulative Distribution Functions of trajectory errors (closest distance between the localization result and the reference trajectory) obtained for different TDOA pairs are presented in Fig.\ref{fig:cdfs}

The proposed adaptive anchor selection method allows to improve localization accuracy compared to using a fixed set of TDOA anchor pairs. In case of the proposed method the median trajectory error was about 25 cm, whereas in case of anchor sets for zones 3 and 4 it was 46 and 50 cm, respectively. In case of the other anchor sets it was not possible to obtain accurate locations in the whole apartment. It might result from unfavorable anchor geometric configuration and problems with signal reception in some areas (zone 5 is located in the kitchen, behind an elevator shaft). 

\section{Conclusions}
\label{sec:conclusions}
The paper presents an adaptive anchor pairs selection method for UWB TDOA-based localization systems. The method divides the area into zones and, based on the calibration performed using a robotic platform, chooses the anchor pairs set, for which the localization RMSE is the lowest.

The simulations have shown that the proposed method improves positioning accuracy and can be used in almost any environment, making it a good alternative to the algorithms presented in the literature.

\bibliographystyle{IEEEtran}
\bibliography{biblio}
%\begin{thebibliography}{00}
%
%\end{thebibliography}
\vspace{12pt}

\end{document}